# Blurred Image Classification based on Adaptive Dictionary

Guangling Sun, Guoqing Li, Jie Yin

School of Communication and Information Engineering, Shanghai University, Shanghai, China

sunguangling@shu.edu.cn

**Abstract**—Two types of framework for blurred image classification based on adaptive dictionary are proposed. Given a blurred image, instead of image deblurring, the semantic category of the image is determined by blur insensitive sparse coefficients calculated depending on an adaptive dictionary. The dictionary is adaptive to the Point Spread Function (PSF) estimated from input blurred image. The PSF is assumed to be space invariant and inferred separately in one framework or updated combining with sparse coefficients calculation in an alternative and iterative algorithm in the other framework. The experiment has evaluated three types of blur, naming defocus blur, simple motion blur and camera shake blur. The experiment results confirm the effectiveness of the proposed frameworks.

**Keywords**- blurred image classification; sparse representation; adaptive dictionary

## I. INTRODUCTION

Image semantic classification remains one of the most challenging problems in computer vision, pattern recognition and statistical learning. To this end, significant progresses have been made in this research area. However, most of the image classification strategies focus on addressing issues such as a wide range of viewpoints, varying scales or illuminations, occlusions and much less attention is devoted to degraded image caused by blur, noise, fog and *etc*. In fact, blur is a very common degradation instance thus recognizing blurred image is significantly meaningful. In this paper, we cope with classifying image degraded by blur in particular.

Compared with general image classification, few literatures exist to handle blurred image classification. Published approaches to this issue can be partitioned into three categories: The first is to extract blur insensitive features. J. Heikkila proposed Local Phase Quantization (LPQ) robust to centrally symmetric blur [1]. In [2], the author declared that the improved LPQ can be applied with any blur regardless of the point spread function. H.Zhang presented orthogonal Lengendre moments to construct a set of invariants to centrally symmetric blur, simple motion blur and noise [3]. The second is to deblur the image followed by classification [4]. M.Nishiyama designed a blurred face recognition framework called FADEIN composed of two stages: first a blur PSF is inferred using Frequency-Magnitude-Based feature space and subspace analysis. Then the deblurred face is recognized based on features used for high quality image recognition. M.Nishiyama further revealed that LPQ extracted from deblurred image actually outperformed comparing with FADEIN or LPQ extracted from blurred image. The third is to make a close combination of image restoration and recognition [5].Although face image is deblurred, the recognition is still accomplished in blur space produced by estimated PSF rather than deblurred space. While these methods are successful to some degree, they all have some limitations. For the first method, the PSF they have tested is simple and has single direction for motion blur. However, some blur such as the camera shake blur are complex and cannot be modeled well with simple motion blur PSF. For the second method, any image deblurring algorithm will inevitably introduce additional artifacts and noise, which in turn have negative affects for classification. As far as the third method, the performance is only evaluated on face recognition application not covering general case.

Our idea belongs to the first category that blur insensitive features are extracted and without image deblurring. Specifically, relying on the framework proposed by Yang in which sparse coefficients of image patch are pooled and used as features to feed and train the SVM classifier [6], sparse coefficients of blurred image patch are directly adopted as features to implement classification without retraining SVM. It is reasonable that sparse coefficients of blurred image patch are regarded as blur insensitive since the used dictionary is adaptive to the specific blur. Hence, learning an adaptive blurred dictionary is a critical component. Certainly, PSF estimation is also a significant important issue. Once the PSF is inferred properly, we can deal with any type of blur resulting from camera defocus, camera shake and simple relative motion between camera and object.

1.1. Overview of proposed framework

As analyzed in above section, dictionary learning is an extremely important in proposed framework. Obviously, sparse coefficients of a blurred image using a sharp dictionary will drift much from that of sharp image. Therefore, we force sharp and blurred image patches to have identical sparse coefficients through dictionary learning to find blur insensitive features.

We proposed two types of frameworks. The first proposed framework is as follows: linear SPM SVM classifier using sparse coefficients of SIFT as features for sharp image is constructed first. Then a PSF is inferred from the input blurred image using method proposed by R. Fergus [7]. Next, the estimated PSF is applied to blur training patches, and SIFT feature of blurred training patches and corresponding sparse coefficients of sharp version are utilized to obtain the adaptive blurred dictionary. Finally, depending on the adaptive dictionary, the sparse coefficients of input image are computed, transformed and utilized to recognize the image. To improve the efficiency and obtain better PSF estimation, the second proposed framework is as follows: joint feature of gradient and SIFT describing a sharp image patch is used to get joint dictionary *D* and sparse coefficients of training images according to *D* act as features to establish SVM classifier. Given an input blurred image, the unknown PSF, the adaptive dictionary and sparse coefficients of patch are updated alternatively and iteratively. The final output sparse coefficients are transformed and utilized to recognize the image. The two types of frameworks are analyzed in detail in section 2.2. Furthermore, to improve computation efficiency, a selection rule is designed to select a small part of all patches to learn the adaptive dictionary as discussed in section 2.3.

## II. PROBLEM FORMULATION

### 2.1 Linear SPM SVM classifier using sparse coefficients

First an image is partitioned into overlapping dense grids and SIFT feature is extracted from each grid. Then SIFT features of all grids are collected together to learn a dictionary and sparse coefficients of each grid are obtained accordingly. Further, three layers of a spatial pyramid for an image is build and each layer is partitioned into $2^l$ parts equally, where *l* denotes *l*th layer, l=0, 1, 2. A 'max' pooling strategy based on sparse coefficient is adopted for each part. Hence all together 21 pooling results are connected as a high dimensional feature vector representing an image. Finally, such feature vectors of training images are utilized to design a linear SVM classifier. The author declared they have achieved states-of-the-art performance [6].

### 2.2 Blurred image classification based on adaptive dictionary

Our work focus on making sparse coefficient of same feature of sharp and blurred patch can be inferred from each other through dictionary learning rather than classifier design so that we directly adopt linear SPM SVM as base classifier in this investigation.

1) Framework I

In fact, there exists an intuitive solution to blurred image recognition: training images are blurred with the PSF estimated from the input blurred image and these blurred training images are further used to learn new classifier. However, it is obvious to be impractical since the classifier needs to be retrained for every unknown image. Thus we propose a trade-off strategy: using a large set of sharp training image patches, a sharp dictionary D and corresponding sparse coefficient matrix $\mathbf{A}_{sh}^{tr}$ are obtained with K-Singular Value Decomposition (KSVD) and Orthogonal Matching Pursuit (OMP) algorithm [8]. Then, a classifier is trained based on D and put aside once training is finished. For any input blurred image, a new dictionary adaptive to the specific PSF inferred from the input image is relearned. Naturally, two essential issues must be addressed: PSF estimation and the adaptive dictionary learning. We adopt Ensemble Learning presented by [7] to infer the PSF. The adaptive dictionary should have a property that sparse coefficient of blurred image patch using it can be utilized to infer that of sharp version using sharp dictionary. To achieve this, we propose to design the following model:

$$\hat{D}_b = \arg\min_{D_b} \left\| P_b^{tr} - D_b \mathbf{A}_{sh}^{tr} \right\|_F^2 \tag{1}$$

Where $P_b^{tr}$ refers to SIFT features extracted from blurred training image patches produced by estimated PSF. Our goal is to search an optimal $\hat{D}_b$ that minimizes the mean approximation errors shown in equation (1). Given a full row rank matrix $\mathbf{A}_{sh}^{tr}$, the solution of this target function can be solved by Method of Optimal Directions (MOD):

$$\hat{D}_b = P_b^{tr} (\mathbf{A}_{sh}^{tr})^T \left[ \mathbf{A}_{sh}^{tr} (\mathbf{A}_{sh}^{tr})^T \right]^{-1} \tag{2}$$

In terms of $\hat{D}_b$ and $\alpha_{sh}^{tr}$, a training patch $p_b^{tr}$ can be approximated as follows:

$$p_b^{tr} \square \hat{D}_b \alpha_{sh}^{tr} = \bar{D}_b \left[ \alpha_{sh,1}^{tr} \square \|\hat{d}_{b,1}\|, \alpha_{sh,2}^{tr} \square \|\hat{d}_{b,2}\|, \cdots, \alpha_{sh,K}^{tr} \square \|\hat{d}_{b,K}\| \right]^T \tag{3}$$

Where $K$ refers to the number of dictionary atom. $\bar{D}_b$ and $\|\hat{d}_{b,j}\|$ denote a normalized dictionary of which each atom is unit vector and $l_2$-norm of $\hat{d}_{b,j}$ respectively. The normalization of $\hat{D}_b$ is a requirement of majority methods of computing sparse coefficient. Moreover, it is assumed that the relation also holds for testing patch:

$$\begin{aligned} p_b^{te} &\square \bar{D}_b \left[ \alpha_{sh,1}^{te} \square \|\hat{d}_{b,1}\|, \alpha_{sh,2}^{te} \square \|\hat{d}_{b,2}\|, \cdots, \alpha_{sh,K}^{te} \square \|\hat{d}_{b,K}\| \right]^T \\ &\square \bar{D}_b \left( \alpha_{b,1}^{te}, \alpha_{b,2}^{te}, \cdots \alpha_{b,K}^{te} \right)^T \end{aligned} \tag{4}$$

It means that during recognition, $\alpha_{sh}^{te}$ could be deduced from $\alpha_b^{te}$ without deblurring the blurred patch.

Certainly, each element of $\boldsymbol{\alpha}_b^{te}$ should be divided by $l_2$-norm of each atom of $\hat{D}_b$.

2) Framework II

As we know, PSF estimation based on Ensemble Learning has intensive time-consuming [7]. To be more efficient and obtain better PSF estimation, we propose another framework making a close combination of PSF estimation and sparse coefficients calculation. In this framework, the scheme of using blur insensitive sparse coefficient for the purpose of recognition is still adopted. Nevertheless, SIFT feature is not appropriate for representing image and meaningless for inferring PSF. To address the issue, we introduce a joint feature of gradient and SIFT as to bridge the gap between recognition and representation. Obviously, the roles of gradient feature have two folds: one is to be used to infer sparse coefficient for recognition and the other is to represent image and estimate PSF. Accordingly, framework II is composed of two phases:

The first phase is to use sharp training images and learn joint dictionary that represent an image patch from two aspects: SIFT feature and gradient feature. A joint dictionary learning model is designed as follows:

$$\{\hat{D}, \hat{\mathbf{A}}^{tr}\} = \arg\min_{D, \mathbf{A}^{tr}} \frac{1}{2} \|P^{tr} - D\mathbf{A}^{tr}\|_F^2 \quad s.t. \quad d_j^T d_j = 1, \|\boldsymbol{\alpha}_i\|_0 \leq L \quad (5)$$
$$i \in \Omega^{tr}, j = 1, 2, \cdots K$$

Where $P^{tr}$ denotes joint data composed of SIFT feature and gradient feature, and $\mathbf{A}^{tr}$ denotes corresponding sparse coefficient. Once $\hat{D}$ is obtained, $D_{grad}$ is to be separated to approximate $p_{grad}^{tr}$ as follows:

$$p_{grad}^{tr} \square D_{grad} \hat{\boldsymbol{\alpha}}^{tr} = \bar{D}_{grad} \left( \hat{\alpha}_1^{tr} \|d_{grad,1}\|, \hat{\alpha}_2^{tr} \|d_{grad,2}\|, \cdots; \hat{\alpha}_K^{tr} \|d_{grad,K}\| \right)^T \quad (6)$$

Similar to expression (3), $\bar{D}_{grad}$ denote a normalized dictionary of which each atom is unit vector. $\hat{D}$ is used to train SVM classifier and $\bar{D}_{grad}$ is utilized to represent image and infer PSF in second phase.

The second phase is to infer PSF and compute sparse coefficient used for recognition.

$$\{\hat{k}, \hat{\mathbf{A}}^{te}\} = \arg\min_{k, \mathbf{A}^{te}} \left\| \nabla B - k \otimes \left[ \sum_{i \in \Omega} R_i^T R_i \right]^{-1} \left[ \sum_{i \in \Omega} R_i^T \bar{D}_{grad} \boldsymbol{\alpha}_i^{te} \right] \right\|_2^2 + \eta \|k\|_2^2, \quad s.t. \|\boldsymbol{\alpha}_i^{te}\|_0 \leq L, i \in \Omega \quad (7)$$

Where $k$, $\nabla B$ and $R_i$ denote PSF, gradient of the input blurred image including horizontal derivative and vertical derivative and a matrix extracting $i$th patch from image respectively. $\|k\|_2^2$ is a Tikhonov regularization term providing a smooth PSF prior and $\eta$ is regularization factor. With Alternating

Minimization scheme, model (7) can be converted into two sub-problems: $k$ estimation and $\mathbf{A}^{te}$ calculation. Before iteratively solving the two sub-problems, one of the two variables must be initialized. We initialize $\mathbf{A}^{te}$ as follows:

$$\hat{\alpha}_i^{te,(0)} = \arg\min_{\alpha_i} \left\| p_{b,i}^{te} - \overline{D}_{grad} \alpha_i^{te} \right\|_2^2, \ s.t. \left\| \alpha_i^{te} \right\|_0 \leq L, i \in \Omega \tag{8}$$

Where $p_{b,i}^{te}$ refers to gradients of $i$th patch of input blurred image. In sequel, two sub-problems are solved alternatively until stop condition is satisfied. We set iteration number as stop condition and usually only very few iteration is required.

a) PSF estimation

Given current $\hat{\mathbf{A}}^{te,(n-1)}$, $k$ is updated to minimize the following model:

$$\begin{aligned} \hat{k}^{(n)} &= \arg\min_k \left\| \nabla B - k \otimes \left[ \sum_{i \in \Omega} R_i^T R_i \right]^{-1} \left[ \sum_{i \in \Omega} R_i^T \overline{D}_{grad} \alpha_i^{te,(n-1)} \right] \right\|_2^2 + \eta \|k\|_2^2 \\ &= \arg\min_k \left\| \nabla B - k \otimes \nabla X \right\|_2^2 + \eta \|k\|_2^2 \end{aligned} \tag{9}$$

$\hat{k}^{(n)}$ is given as follows:

$$\hat{k}^{(n)} = F^{-1}\left( \frac{\overline{F(\partial_x X)}F(\partial_x B) + \overline{F(\partial_y X)}F(\partial_y B)}{\overline{F(\partial_x X)}F(\partial_x X) + \overline{F(\partial_y X)}F(\partial_y X) + \eta \mathbf{I}} \right) \tag{10}$$

Where $F(\cdot)$ and $F^{-1}(\cdot)$ denote the FFT and inverse FFT respectively. $\overline{F(\cdot)}$ is a complex conjugate operator.

b) Sparse Coefficients calculation

Given current $\hat{k}^{(n)}$, $\mathbf{A}^{te}$ is updated in following ways: first, $\hat{k}^{(n)}$ is used to blur all training image patches and adaptive $\hat{D}_{b,grad}^{(n)}$ is obtained similar to expression (2); then $\hat{\mathbf{A}}^{te,(n)}$ is updated as follows:

$$\hat{\alpha}_{b,i}^{te,(n)} = \arg\min_{\alpha_{b,i}^{te}} \left\| p_{b,i}^{te} - \overline{D}_{b,grad}^{(n)} \alpha_{b,i}^{te} \right\|_2^2, \ s.t. \left\| \alpha_{b,i}^{te} \right\|_0 \leq L, i \in \Omega \tag{11}$$

$$\hat{\alpha}_i^{te,(n)} = \left( \hat{\alpha}_{b,i,1}^{te,(n)} / \left\| \hat{d}_{b,grad,1}^{(n)} \right\|, \hat{\alpha}_{b,i,2}^{te,(n)} / \left\| \hat{d}_{b,grad,2}^{(n)} \right\|, \cdots \hat{\alpha}_{b,i,K}^{te,(n)} / \left\| \hat{d}_{b,grad,K}^{(n)} \right\| \right)^T \tag{12}$$

Where $\overline{D}_{b,grad}^{(n)}$ refers to normalized blurred dictionary after $n$th iteration. Once preset iteration number $T$ is reached, final sparse coefficients used for recognition are obtained according to expression (6) as follows:

$$\hat{\boldsymbol{\alpha}}_i = \left( \hat{\alpha}_{i,1}^{\prime e,(T)} / \|d_{grad,1}\|, \hat{\alpha}_{i,2}^{\prime e,(T)} / \|d_{grad,2}\|, \cdots \hat{\alpha}_{i,K}^{\prime e,(T)} / \|d_{grad,K}\| \right)^T \quad (13)$$

In sum, for gradient feature and blurred and sharp patch, the sparse coefficients of them using $\hat{D}_{b,grad}^{(n)}$ and $\bar{D}_{grad}$ are related by a set of factors; meanwhile, for joint feature composed of SIFT and gradient and gradient alone, the sparse coefficients of them using $\hat{D}$ and $\bar{D}_{grad}$ are also related by a set of factors. Hence, sparse coefficient of gradient feature of a blurred patch obtained from (11) can be utilized to predict that of joint feature of its sharp version as described in (12) and (13).

2.3 Efficiency improvement consideration

In both the two frameworks, a large set of training patches is used to get dictionary for classifier design. However, for adaptive blurred dictionary learning, it is not necessary to utilize all the training patches. On the other hand, it is well known that only support vectors are needed using SVM to classify a pattern, thus it is reasonable to assume the support vectors contain most of the useful information for recognition. Thereby, to achieve a good trade-off between efficiency and performance, we propose an acceleration scheme: only a part of the large set of training patches coming from the training images corresponding to support vector images are blurred and utilized to learn the adaptive blurred dictionary.

III. EXPERIMENT AND RESULT

We implement the proposed frameworks and carry out experiments on Matlab platform.

3.1 Image database

The tested database is Caltech101 and all 102 categories are trained. Random 20 images of each category are selected as training samples. Altogether 10 categories including accordion, pizza, buddha, car-side, leopards, lotus, pyramid, rooster, gramophone and Windsor-chair. Among these 10 categories, 20 samples of each category are tested to evaluate the proposed framework. Some samples of Caltech101 have been listed in figure 1.

3.2 Tested blue kernel

The tested blur kernels (PSF) are Gaussian kernel, motion kernel 1 generated by Matlab function and motion kernel 2 provided by Levin [9] respectively. The details of them are listed in following:

(1) Gaussian kernel: Gaussian low pass filter with size 9*9 and standard deviation 5. The kernel simulates blur resulting from camera defocus.

(2) Motion kernel 1: Linear motion of 20 pixels length and direction 45o.The kernel simulates blur resulting from simple relative motion between object and camera.

(3) Motion kernel 2: The sixth kernel chosen from file: LevinEtalCVPR09Data.rar. The kernel simulates blur resulting from camera shake.

The three kernels and corresponding blurred images have been illustrated in figure 2.

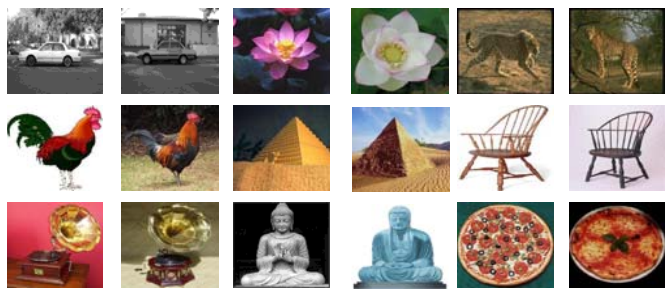

Fig. 1. Sample Images in Caltech101

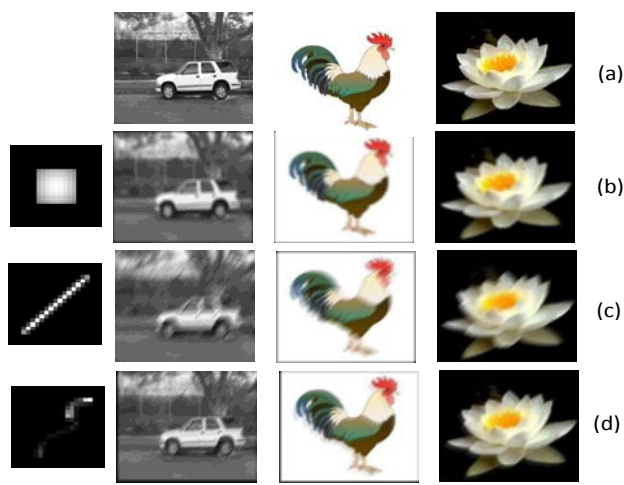

Fig. 2. (a) Sharp images. (b) Gaussian kernel and blurred images. (c) Motion kernel 1 and blurred images. (d) Motion kernel 2 and blurred images.

3.3 Algorithm and classifier parameter setting

All parameters related with SVM classifier in two frameworks are the same as in Ref [6]. Sparse degrees are all set as $L$=5 and number of dictionary atom is set as $K$=1024. In framework II, $\eta$ and $T$ are set as $\eta = 4$ and $T = 5$ respectively. The size of grid is selected as 16*16 pixels and the dimensionality of gradient feature is 512 accordingly. Consequently, Principal Component Analysis technique is used to reduce the dimensionality before combining with SIFT feature.

Altogether four methods and three kernels are evaluated in the experiment and the result is listed in table I. Besides the proposed two frameworks, other two compared methods are: one is to recognize with sharp dictionary; the other is to deblur the blurred image with Richardson-Lucy algorithm before recognition. The recognition accuracy of sharp image is 75% which roughly agree with the result reported by [6]. However, for recognizing blurred image that still uses sharp dictionary, the performance declined dramatically. After removing blurring with Richardson-Lucy algorithm, accuracy has increased to some degree. But the proposed frameworks have obtained higher accuracy and especially, the highest accuracies have been achieved by framework II for three blur kernels.

TABLE I

ACCURACY COMPARISON OF MULTI-METHODS AND MULTI-KERNELS

| Method \ Kernel | Using sharp dictionary | Deblurring with R-L algorithm | Framework I | Framework II |
|---|---|---|---|---|
| Gaussian | 47.5% | 61% | 64% | **67.5%** |
| Motion1 | 46.5% | 57% | 62% | **66%** |
| Motion2 | 55.5% | 71% | 69% | **72%** |
| The accuracy of sharp image classification is 75% | | | | |

IV. CONCLUSIONS AND FUTURE WORK

In this paper, we propose two types of framework for blurred image classification and space-invariant blur kernel is assumed. The two frameworks are based on adaptive dictionary and neither demands image deblurring. The essential idea is that a new dictionary being capable of adaptive to inferred PSF from input blurred image is relearned for every input image. Therefore, for each blurred image patch, the sparse coefficient obtained by adaptive dictionary is insensitive to arbitrary blur. Meanwhile, for the two frameworks, the performance of the latter is higher than that of the former, since the former infers the PSF as a separate step, and the latter updates the PSF and sparse coefficient of gradient feature alternatively so as to better combine PSF estimation and sparse coefficient calculation. The proposed framework can tackle any blur resulting from camera defocus, simple relative motion between camera and object, to camera shake.

The further work may come from two aspects: one is adaptive dictionary learning will not rely on outer image database but itself; the other is to cope with space variant blurred image recognition.